\documentclass[preprint]{elsarticle}

\usepackage{hyperref}
\usepackage[ruled,vlined]{algorithm2e}
\SetArgSty{textnormal}

\usepackage{siunitx}
\usepackage{amsthm}
\usepackage{amssymb}
\usepackage{amsmath}

\theoremstyle{definition}
\newtheorem{definition}{Definition}

\journal{Information Systems}

\begin{document}

\begin{frontmatter}

\title{Local Intrinsic Dimensionality Measures for Graphs, with Applications to Graph Embeddings\tnoteref{mytitlenote}}
\tnotetext[mytitlenote]{This is an extended version of our conference paper~\cite{2021_NCLID_SISAP} presented at SISAP 2021 -- {\it 14th International Conference on Similarity Search and Applications}.}

\author[ouraddress]{Milo\v{s} Savi\'{c}\corref{mycorrespondingauthor}}
\cortext[mycorrespondingauthor]{Corresponding author}
\ead{svc@dmi.uns.ac.rs}

\author[ouraddress]{Vladimir Kurbalija}
\ead{kurba@dmi.uns.ac.rs}

\author[ouraddress]{Milo\v{s} Radovanovi\'c}
\ead{radacha@dmi.uns.ac.rs}

\address[ouraddress]{University of Novi Sad, Faculty of Sciences, Department of Mathematics and Informatics\\Trg Dositeja Obradovi\'ca 4, 21000 Novi Sad, Serbia}

\begin{abstract}
The notion of local intrinsic dimensionality (LID) is an important advancement 
in data dimensionality analysis, with applications in data mining, 
machine learning and similarity search problems. Existing distance-based
LID estimators were designed for tabular datasets encompassing 
data points represented as vectors in a Euclidean space.
After discussing their limitations for graph-structured data 
considering graph embeddings and graph distances, we propose NC-LID, a novel LID-related measure for 
quantifying the discriminatory power of the shortest-path distance with respect to natural communities 
of nodes as their intrinsic localities.
It is shown how this measure can be used to design LID-aware graph embedding algorithms
by formulating two LID-elastic variants of node2vec with personalized hyperparameters that 
are adjusted according to NC-LID values.
Our empirical analysis of NC-LID on a large number of real-world graphs shows 
that this measure is able to point to nodes with high link reconstruction errors 
in node2vec embeddings better than node centrality metrics. The experimental evaluation also 
shows that the proposed LID-elastic node2vec extensions improve node2vec 
by better preserving graph structure in generated embeddings. 
\end{abstract}

\begin{keyword}
intrinsic dimensionality \sep graph embeddings \sep graph distances \sep natural communities \sep LID-elastic node2vec
\end{keyword}

\end{frontmatter}


\section{Introduction}

The intrinsic dimensionality (ID) of a dataset is the minimal number of features
that are needed to form a good representation of data in a lower-dimensional
space without a large information loss. Estimation of ID is highly relevant 
when performing dimensionality reduction in various machine learning (ML) and data mining (DM) tasks.
Namely, machine learning models can be trained on lower-dimensional data representations 
in order to achieve a higher level of generalizability by alleviating
negative impacts of high dimensionality (e.g., negative phenomena associated with 
the curse of dimensionality). Due to a smaller number of features,
such models are additionally more comprehensible, while their training, tuning 
and validation is more time efficient. 

The notion of local intrinsic dimensionality (LID) has been developed in recent years
motivated by the fact that the ID may vary across a dataset. The main idea of LID
is to focus the estimation of ID to a data space surrounding a data point.
In a seminal paper, Houle~\cite{Houle2013} defined the LID considering the distribution of 
distances to a reference data point. Intuitively, the LID of the reference data point
expresses the degree of difficulty to separate its nearest neighbors from the rest
of the dataset. Houle showed that for continuous distance distributions with differentiable cumulative 
density functions the LID and the indiscriminability of the corresponding distance function 
are actually equivalent. 

Inspired by the works of Houle and collaborators~\cite{Houle2013,Amsaleg2015}, in~\cite{2021_NCLID_SISAP} we proposed  
an approach for formulating LID-related measures for nodes in graphs and complex networks. By using 
the proposed approach we defined one concrete LID measure called NC-LID that is based on natural or local 
communities~\cite{Lancichinetti_2009} as intrinsic subgraph localities. In~\cite{2021_NCLID_SISAP} we also 
designed two LID-elastic node2vec~\cite{nodetovec2016} extensions based on NC-LID measure. 
This works extends~\cite{2021_NCLID_SISAP} in the following ways:
\begin{enumerate}
 \item The empirical evalution of NC-LID and LID-elastic no\-de\-2\-vec extensions is expanded from 
 10 to 18 real-world networks (15 of them being large-scale). The obtained results confirm the initial findings 
 from~\cite{2021_NCLID_SISAP} that (1) NC-LID significantly correlates to link reconstruction errors 
 when reconstructing input graphs from node2vec embeddings, and (2) LID-elastic no\-de\-2\-vec extensions outperform 
 node2vec by providing graph embeddings that better preserve the structure of the input graphs.
 \item We empirically compare NC-LID to various node centrality metrics quantifying structural importance of nodes
 in complex networks regarding their ability to identify nodes with high link reconstruction errors in 
 node2vec embeddings. The results of this analysis show that NC-LID has consistent and stronger correlations to link reconstruction 
 errors than existing node centrality measures. This finding implies that our measure is a better choice for 
 designing ``elastic'' graph embedding algorithms in which hyper-parameters are adjusted according 
 to structural measures associated to nodes. 
\end{enumerate}
Additionally, subsequent research by Kne\v{z}evi\'{c} et al.~\cite{Knezevic22} showed that our 
NC-LID based no\-de\-2\-vec extensions outperform no\-de\-2\-vec when utilizing generated embeddings 
in a concrete ML/DM application (i.e., node clustering). The implementation of NC-LID and  
NC-LID based node2vec extensions together with evaluation procedures 
and datasets used in this paper are contained in the open-source GRASPE 
library\footnote{The GRASPE library is publicly available at https://github.com/graphsinspace/graspe.}.

The rest of the paper is structured as follows. After presenting relevant research works in Section~\ref{RW},
the motivation for this work and its main contributions are outlined in Section~\ref{MC}. The NC-LID measure
is discussed in Section~\ref{nclid_measure}, while the LID-elastic node2vec extensions based on NC-LID are explained in 
Section~\ref{lid_elastic_n2v}. The experimental evaluation of NC-LID and NC-LID based no\-de\-2\-vec extensions are presented in Section~\ref{experiments_results}. The last section concludes the paper and discusses possible directions for future research.

\section{Related Work}
\label{RW}

\subsection{Local Intrinsic Dimensionality}
\label{RW_LID}

The theoretical foundation of LID was set by Houle who formally defined it~\cite{Houle2013}
and then explored its mathematical properties in a series of papers~\cite{Houle_LID1,Houle_LID2,Houle2020_LID3}.
Let $x$ be a reference data point and let $F$ denote the cumulative
distribution function of distances to $x$. It can be said that the underlying distance function 
is discriminative 
at a given distance $r$ if $F(r)$ has a small increase for a small
increase in $r$. Thus, the indiscriminability of the distance function at $r$ w.r.t $x$,
denoted by $\mbox{Ind}(r)$,
can be quantified as: 
\begin{equation*}
\mbox{Ind}(r) = \lim_{\varepsilon \rightarrow 0} {\frac{F((1 + \varepsilon)r) - F(r)}{\varepsilon \: F(r)}}. 
\end{equation*}
Following the generalized expansion model (GED)~\cite{GED2012}, Houle defined the local intrinsic dimensionality of $F$ at $r$,
denoted by $\mbox{IntrDim}(r)$, by substituting ball volumes in GED by distance probability 
measures:
\begin{equation*}
\mbox{IntrDim}(r) = \lim_{\varepsilon \rightarrow 0} \frac{\mbox{ln}F((1 + \varepsilon)r) - \mbox{ln}F(r)}{\mbox{ln}(1 + \varepsilon)}. 
\end{equation*}
In~\cite{Houle2013} it is shown that Ind and IntrDim are actually equivalent, both representing the intrinsic 
dimensionality at distance $r$, denoted by $\mbox{ID}(r)$:
\begin{equation*}
\mbox{IntrDim}(r) = \mbox{Ind}(r) \triangleq \mbox{ID}(r). 
\end{equation*}
Then, the LID of $x$ can be defined when $r$ goes to 0 as $\lim_{r \to 0} \mbox{ID}(r)$.

For practical applications, the LID of $x$ can be estimated
considering the distances of $x$ to its $k$ nearest data points~\cite{Amsaleg2015,Amsaleg2019}.
In~\cite{Amsaleg2015}, Amsaleg et al. proposed LID estimators based on maximum-likelihood estimation (MLE),
the method of moments, the method of probability-weighted moments and regularly varying functions. In particular,
the MLE estimator defined by the following expression
\begin{equation*}
\mbox{MLE-LID}(x) = -\left(\frac{1}{k}\sum_{i=1}^{k}\ln\frac{x_{i}}{x_{k}}\right)^{-1}
\end{equation*}
provides a good trade-off between complexity and statistical accuracy 
($x_{i}$ is the distance between $x$ and its $i$-th nearest neighbour). In~\cite{Amsaleg2019}, 
the authors extended the previous work by proposing a MLE-based LID estimator suitable for tight localities, 
i.e. neighborhoods of small size that are extremely important in outlier detection and
nearest-neighbors classification.

The development of the LID model is in large motivated by similarity search problems. It was shown that 
pruning operations and early termination 
of similarity search queries can be improved by LID-aware methods~\cite{Casanova2017,Houle2014_subspace,Houle2012_DimensionalityTesting}. 
The LID can also improve the accuracy of methods constructing approximate kNN graphs~\cite{Houle2017_KNN} 
that are relevant not only to similarity search problems, but also for distance-based classification methods.
Additionally, LID can be exploited for benchmarking nearest neighbors search~\cite{AUMULLER2021}.
Based on the LID model, von Ritter et al.~\cite{vonRitter2018} derive an estimator of the local growth rate of 
the neighborhood size for similarity search on graphs. 
LID-aware methods were proposed also in other fields of machine learning and data mining, e.g., in outlier
detection~\cite{Houle2018_Outliers} and subspace clustering with estimation of local relevance 
of features~\cite{becker2019subspace}.
Existing research works also indicate that LID can be applied in deep learning. 
Ma et al.~\cite{Ma2018_noisyLabels} proposed a LID-aware method for training deep neural 
network classifiers on datasets with noisy labels, while the authors of~\cite{ma2018characterizing} demonstrate
that LID can be used to detect adversarial data points when training deep neural networks.

\subsection{Graph Embeddings}

Graphs are dimensionless objects. The main purpose of graph embedding algorithms is to project a graph into
an Euclidean space of a given dimension such that the structure of the graph is well preserved with respect
to shortest-path distances among nodes. In this way, machine learning algorithms designed for tabular data 
can be applied to graphs, thus enabling various applications such as node classification, node clustering and
link prediction, to mention a few of the most important. Due to a ubiquitous presence of graphs and networks in various
domains, the field of graph embedding algorithms is currently a very active research area. 
For good overviews of the most important methods we refer readers to articles by Goyal and Ferrara~\cite{GOYAL2018},
Cai et al.~\cite{Cai2018} and Makarov et al.~\cite{Makarov2021SurveyOG}.

The first graph embedding algorithms proposed in the literature were based on matrix factorization approaches
used in dimensionality reduction techniques. With the rise of representational learning methods, the focus of 
researchers shifted to methods based on random walks and deep learning techniques, which are currently two dominant 
categories of graph embedding algorithms.

The principal idea of random walk methods is to sample a certain number of random walks for each node in order to
capture its neighborhood. The sampled random walks are then treated as sentences composed of node identifiers and 
the problem of generating graph embeddings is reduced to the problem of generating word embeddings.
The sampling of random walks can be unbiased (i.e., each neighbor has an equal probability to be visited
in the next random walk step) such as in DeepWalk~\cite{deepwalk2014} or it can be based on some biased 
walking strategies such as in node2vec~\cite{nodetovec2016}. Random walking strategies could also be designed
to preserve some higher-order graph properties such as structural roles~\cite{struc2vec}.

Graph embedding methods using deep learning techniques form embeddings by training neural networks 
and then exploting learned latent representations encoded in their middle layers. In unsupervised settings, 
trained neural networks are autoencoders preserving adjacency matrices~\cite{Wang2016} or 
matrices encoding higher-order similarities~\cite{Cao2016}. On the other hand, supervised and semi-supervised
graph learning algorithms such as graph convolutional networks~\cite{Kipf2017} 
and graph aggregation networks~\cite{Hamilton2017}, 
can be applied to obtain embeddings of labeled and/or attributed graphs.

\section{Motivation and Contributions}
\label{MC}

Existing distance-based LID estimators have been designed for
tabular data\-sets with real-valued features and smooth distance functions.
There are two ways in which they can be applied to graphs:
\begin{enumerate}
\item by transforming graphs into tabular data representations using graph embedding algorithms, and
\item by using graph-based distances instead of distances of vectors in Euclidean spaces.
\end{enumerate}

The first approach enables LID-based assessments of graph embeddings
and their analysis in the context of distance-based machine learning 
and data mining algorithms. For example, the maximum-likelihood LID estimator 
(MLE-LID, see Section~\ref{RW_LID}) proposed by Amsaleg et al.~\cite{Amsaleg2015} can be computed on 
graph embeddings produced by different graph embedding methods. In this way
we can determine which of the methods is the most effective
for distance-based machine learning and data mining algorithms
if the produced embeddings preserve the structure 
of the graph to a similar extent. Furthermore, obtained MLE-LID values 
and their distributions can indicate whether we can benefit from LID-aware data mining 
and machine learning algorithms.

It is important to emphasize that graph embedding algorithms produce embeddings of an explicitly
specified dimensionality. Consequently, LID estimates for graph nodes obtained via embeddings
are relative to the selected embedding dimensionality. Additionally, the accuracy 
of obtained LID estimates depends on the ability of the selected graph embedding algorithm 
to preserve the structure of the input graph.

In the most general case, the goal of graph embedding algorithms is to
preserve shortest path distances in generated embeddings. Therefore, 
existing LID estimators can be applied ``directly'' on graphs
by taking shortest path distances instead of distances in Euclidean spaces.
However, LID estimates based on shortest path distances will suffer from 
negative effects of the small-world property~\cite{watts_collective_1998}, i.e. for a randomly selected node 
$n$ there will be an extremely large fraction of nodes at the same 
and relatively small shortest-path distance from $n$. 
The scale-free property of large-scale real-world graphs~\cite{Barabasi99emergenceScaling}
(i.e., the existence of nodes with an extremely high degree that also called hubs) 
will also have a big impact on such LID estimates.
For example, LID for hubs will be estimated as 0 by the MLE-LID estimator due to
a large number of nearest neighbors at the shortest-path distance 1.
Another problem with this approach is the shortest-path distance itself. 
The notion of LID is based on the
assumption that the radius of a ball around a data point can be increased by
a small value that tends to 0. However, the shortest-path distance does
not have an increase that can go to 0 (the minimal increase is 1)
in contrast to distances in Euclidean space. Alternatively, graph spectral
analysis (related to eigenvector-based node centrality measures) could be 
used to approximate the LID of graph nodes. For example, it is possible to derive 
the complexity of local structures in tabular datasets by 
diffusions~\cite{COIFMAN2006} or random walks~\cite{Maillet2021} over neighborhood graphs (e.g., kNN
graphs).

Considering the previous discussion, we take a different approach to designing LID measures
for nodes in a graph. The main idea of our approach is to substitute
a ball around a data point with a subgraph around a node in order to estimate the discriminatory
power of a graph-based distance of interest. We consider the most basic case in which a fixed
subgraph is taken as the intrinsic locality of the node. Our first contribution is the definition
of a general form of the graph-based LID that reflects the local degree of the discriminatory 
power of an arbitrary graph-based distance function. From this general form one concrete
measure called NC-LID is devised by taking shorest-path distance as the underlying distance function and 
natural or local communities of nodes as their intrinsic localities. 

Our empirical evaluation of NC-LID on a large number of real-world graphs indicates that
NC-LID is able to identify nodes with high link reconstruction errors in node2vec
embeddings. It is important to mention that node2vec was selected
as the baseline graph embedding method to evaluate NC-LID after we tuned 
and examined several state-of-the-art graph embedding algorithms also including methods based on graph neural networks. 
On real-world graphs from our experimental corpus, node2vec preserves the structure of embedded graphs to the best 
extent, i.e.\ it tends to have the lowest graph reconstruction errors among 4 considered alternatives: 
DeepWalk~\cite{deepwalk2014}, graph convolutional networks~\cite{Kipf2017}, graph autoencoders~\cite{Wang2016} 
and GraphSAGE~\cite{Hamilton2017}. 
More specifically, node2vec does not achieve the highest $F_{1}$ graph reconstruction score
(for the definition of the $F_{1}$ score please see Section~\ref{lid_node2vec_eval}) 
only for one graph (CITESEER) from our corpus: for this graph the highest $F_{1}$ is obtained by the 
graph autoencoder, while node2vec gives the second highest $F_{1}$.

As the second contribution of the paper, we demonstrate that NC-LID is a better indicator of nodes with
high link reconstruction errors in node2vec embeddings than various existing node centrality measures 
(degree, betweenness, closeness, eigenvector centrality and core index). 
This result implies that NC-LID is a better choice for designing elastic graph embedding algorithms in which 
hyper-parameters are personalized per node and/or link, and adjusted from some base values according 
to structural graph measures.
Finally, our empirical evaluation of the proposed LID-elastic node2vec extensions on a large set of 
real-world graphs shows that the NC-LID measure can effectively improve the quality 
of node2vec embeddings.


\section{NC-LID Measure}
\label{nclid_measure}

Let $n$ denote a node in a graph $G = (V, E)$, let $S$ be a subgraph containing $n$
and let $d$ be a graph-based distance of interest. The distance $d$ can be the 
shortest-path distance, but also any other node similarity 
function~\cite{Savic2019Fundametal},
including hybrid node similarity measures for attributed graphs 
(a measure combining graph-based
similarity of nodes with similarity of their attributes).
Assuming that $S$ is a natural (intrinsic) locality of $n$, $d$ can be considered as a perfectly
discriminative distance measure if it clearly separates nodes in $S$ from the rest of the
nodes in $G$, i.e.
\begin{equation}
(\forall s \in S) \: (\forall r \in V \setminus S) \: \: d(n, s) < d(n, r),
\end{equation}
or, equivalently,
\begin{equation}
\label{second_form}
\max_{s\in S} d(n,s) < \min_{r \not \in S} d(n,r).
\end{equation}
Equation~\ref{second_form} implies that a graph embedding function $f$ mapping nodes to 
points in a Euclidean space should be non-expansive, which in turn relates approximate 
search in Euclidean spaces to the LID of graph nodes.

To quantify the degree of discriminatory power of $d$ considering $S$ as the intrinsic
locality of $n$, we define a general limiting form of graph-based local intrinsic
discriminability (GB-LID) as
\begin{equation}
\mbox{GB-LID}(n) = - \ln \left( \frac{|S|}{T(n, S)} \right),
\end{equation}
where $|S|$ is the number of nodes in $S$. $T(n, S)$ is the number of nodes whose
distance from $n$ is smaller than or equal to $\rho$, where $\rho$ is the maximal distance
between $n$ and any node from $S$:
\begin{equation}
\label{T_equation}
T(n, S) = \left| \left\{ y \in V \: : \: d(n, y) \leq \max_{z \in S} \: d(n, z) \right\} \right|.
\end{equation}

Similarly to the standard LID for tabular data, GB-LID assesses the local neighborhood size
of $n$ at two ranges: 
\begin{enumerate}
 \item the number of nodes in a neighborhood of interest ($S$), and
 \item the total number of nodes that are located from $n$ within a relevant radius 
 (the maximal distance from $n$ to any node in $S$).
\end{enumerate}
The more extreme the ratio between these two quantities, 
the higher complexity of $S$ and the local intrinsic dimensionality of $n$. 
Unlike standard LID, GB-LID depends on the complexity of a fixed subgraph around the node
rather than some measure reflecting the dynamics of expanding subgraphs 
(LID measures based on expanding subgraphs will be part of
our future research). 
Compared to other measures capturing the local complexity of a node
(e.g., degree centrality), GB-LID is not restricted to ego-networks
of nodes or regularly expanding subgraphs capturing all nodes within 
the given distance (e.g., LID-based intrinsic degree proposed 
by von Ritter et al.~\cite{vonRitter2018}).

GB-LID is a class of LID-related scores effectively parameterized
by $S_{n}$ and $d$, where $S_{n}$ is the subgraph denoting
the intrinsic local neighborhood of node $n$ and $d$ is an underlying
distance measure. From GB-LID we derive one concrete measure called NC-LID
(NC is the abbreviation for ``Natural Community''). In NC-LID we fix $S_{n}$
to the natural (local) community of $n$ determined by the fitness-based
algorithm for recovering natural communities~\cite{Lancichinetti_2009} and $d$ is
the shortest path distance. After identifying the natural community of $n$,
NC-LID for $n$ can be computed by a simple BFS-like algorithm (see Algorithm 1).


%
%
%
%
%

\begin{algorithm}[htb!]
\small
\SetAlgoLined
\DontPrintSemicolon
\SetKwInOut{Input}{input}
\SetKwInOut{Output}{output}
\Input{$G$ -- a graph, $n$ -- a node in $G$}
\Output{NC-LID of $n$} 

\BlankLine \BlankLine 
$S$ = identify the natural community of $n$ by the algorithm proposed by Lancichinetti et al.~\cite{Lancichinetti_2009}\\
$M = \max_{z \in S}$ shortest-path-distance($n$, $z$) \\
$Q$ = an empty queue of nodes\\
append $n$ to $Q$ \\
mark $n$ as visited \\
\BlankLine

// initialize the number of nodes in $T(n, S)$ (defined by equation~\ref{T_equation}) to 0 \\
$T = 0$ \\ 

\BlankLine \BlankLine

\While{$Q$ is not empty} {
    $c$ = remove the first element from $Q$ \\
    $T = T + 1$ \\
    $d$ = shortest-path-distance($n$, $c$) \\
    \uIf{$d > M$} {
       break \\
    }\Else{
        $P$ = retrieve nearest-neighbors of $c$ in $G$ \\
        \ForEach {non-visited node $p$ in $P$} {
            mark $p$ as visited \\
            append $p$ to $Q$
        }
    }
}

\BlankLine

{\bf return} $-\ln(|S| / T)$ 

\caption{{\bf The algorithm for computing NC-LID}}	
\SetAlgoRefName{Al1}
\SetAlgoCaptionSeparator{'.'}
\end{algorithm}

A community in a graph is a highly cohesive subgraph~\cite{FORTUNATO2010}. 
This means that the number of links within the community (so-called intra-community links)
is significantly higher than the number of links connecting nodes from the
community to nodes outside the community (so-called inter-community links).
The notion of community can be trivially expanded to weighted graphs by taking
link weights instead of their counts.
The natural or local community of node $n$ is a community
recovered from $n$~\cite{Clauset2005,Bagrow2008}. Algorithms for 
identifying natural communities maintain two sets of nodes:
the natural community $C$ and the border set of nodes adjacent to $C$
denoted by $B$. Typically, one or more nodes from $B$
are selected to expand $C$, then $B$ is updated to include any new 
discovered neighbors. The previous operation continues until an
appropriate stopping criterion has been met.

Natural communities used in the NC-LID measure are identified by the 
fitness-based algorithm proposed by Lancichinetti et al.~\cite{Lancichinetti_2009}.
This algorithm recovers the natural community $C$ of
$n$ by maximizing the community fitness function that is defined as:
\begin{equation}
f_{C} = \frac{k_{in}(C)}{(k_{in}(C) + k_{out}(C))^{\alpha}},
\end{equation}
where $k_{in}(C)$ is the total intra-degree of nodes in $C$,
$k_{out}(C)$ is the total inter-degree of nodes in $C$, and
$\alpha$ is a real-valued parameter controlling the size of
$C$ (larger $\alpha$ implies smaller $C$).
The intra-degree and inter-degree of a node $s$ are the number of
intra-community and inter-community links incident to $s$, respectively.
The most natural choice for $\alpha$ is $\alpha = 1$, which corresponds
to the Raddichi notion of weak communities~\cite{Radicchi2004}. 
The realization of the algorithm is based on the concept of node fitness
that is defined as the difference of $f_{C}$ with and without some concrete 
node.
In each iteration, the algorithm performs the following steps:
\begin{enumerate}
 \item The node with largest positive fitness from the border $B$ is added to $C$.
 \item The fitness of each node is recalculated after expanding $C$ and nodes
 with negative fitness are excluded from $C$.
 \item The previous step is repeated until there are no nodes in $C$ having negative
 fitness. Otherwise, the algorithm starts the next iteration from the
 first step.
\end{enumerate}
The algorithm stops when all nodes from $B$ have negative fitness. 
It is also important to observe that the algorithm for detecting
natural communities is a deterministic procedure that is not dependent
on the order of nodes in adjacency lists: (1) the node from $B$
maximally increasing the fitness function is selected to expand $C$, and
(2) all nodes in $B$ with negative fitness are removed from $C$ prior to
recalculating fitness of nodes in $C$.

NC-LID($n$) is equal to 0 if all nodes from the natural community of $n$
are at smaller shortest-path distances to $n$ than nodes outside its
natural community. Higher values of NC-LID($n$) imply that it is harder to
distinguish the natural community of $n$ from the rest of the graph using
the shortest-path distance, i.e. the natural community of $n$ tends
to be more ``concave'' and elongated in depth with higher NC-LID($n$)
values. 

\section{LID-elastic node2vec}
\label{lid_elastic_n2v}

LID-based measures for graph nodes, such as NC-LID introduced in the previous 
section, enables us to design LID-aware or LID-elastic graph embedding algorithms.
We propose two LID-elastic variants of node2vec~\cite{nodetovec2016} in which node2vec
hyperparameters are personalized at the level of nodes and/or links and adjusted
according to NC-LID values.

Node2vec is a random-walk based algorithm for generating graph embeddings.
A certain fixed number of random walks is sampled from each node in a graph.
Sampled random walks are then treated as text sentences in which
node identifiers are tokens (words). 
Node2ec uses word2vec skip-gram architecture~\cite{Mikolov2013} 
to make vectors of node embeddings from random-walk sentences.

Node2vec is based on a biased random walk sampling strategy. The sampling
strategy is controlled by two parameters: the return parameter $p$ and 
the in-out parameter $q$. 
Let us assume that a random walk just transitioned from node $t$ to node $v$ and
let $x$ be a neighbour of $v$. The unnormalized probability of transitioning from 
$v$ to $x$ is given by 
\begin{equation}
\label{transition_probabilities}
Pr(v \rightarrow x) = 
\begin{cases}
1/p       & \mbox{if distance}(t, x) = 0 \\
1   & \mbox{if distance}(t, x) = 1\\
1/q & \mbox{if distance}(t, x) = 2
\end{cases}
\end{equation}
It can be seen that parameter $p$ determines the probability 
of intermediately returning back to $t$.
Parameter $q$ controls to what extent the random walk resembles
BFS or DFS graph exploration strategies. For $q > 1$, the random walk is more biased
to nodes close to $t$ (BFS-like graph exploration). If $q < 1$ then
the random walk is more inclined to visit nodes that are further away from $t$ 
(DFS-like graph exploration).

The mechanism for sampling random walks in node2vec is controlled by 4 hyperparameters:
the number of random walks sampled per node, the length of each random walk,
the return parameter $p$ and the in-out parameter $q$. The first two parameters are fixed for each node in the graph, while
$p$ and $q$ are fixed for each link.
Our LID-elastic node2vec extensions are based on ``elastic'', non-fixed node2vec hyperparameters 
that are customized for each node and link according to NC-LID values.

The first LID-elastic variant of node2vec, denoted by \verb|lid-n2v-rw|, personalizes
only node-related hyperparameters that are adjusted from some fixed base values. 
The pseudo-code for \verb|lid-n2v-rw| is given in Algorithm 2.
The number of random walks sampled for a node $v$ 
is equal to
\begin{equation}
\label{NRW}
\mbox{NRW}(v) = \lfloor (1 + \mbox{NC-LID}(v)) \cdot B \rfloor, 
\end{equation}
where $B$ is the base number of random walks (by default $B = 10$). The length of each sampled 
walk starting from $v$ is determined by the following equation:
\begin{equation}
\label{LRW}
\mbox{LRW}(v) = \lfloor W / (1 + \mbox{NC-LID}(v)) \rfloor,
\end{equation}
where $W$ denotes the base random walk length (by default $W = 80$).

\begin{algorithm}[htb!]
\small
\SetAlgoLined
\DontPrintSemicolon
\SetKwInOut{Input}{input}
\SetKwInOut{Output}{output}
\Input{$G$ -- a graph}
\Output{$E$ -- an embedding of $G$} 

\BlankLine \BlankLine 
$R$ = empty list of sampled random walks \\
$V$ = set of nodes in $G$ \\
\ForEach{$v \in V$} {
    // determine the number of random walks that will be sampled for $v$ \\
    $n = \mbox{NRW}(v)$ (Equation~\ref{NRW})\\
    \BlankLine
    
    // determine the length of random walks sampled for $v$ \\
    $l = \mbox{LRW}(v)$ (Equation~\ref{LRW})\\
    \BlankLine
    
    \For{$i = 1$ \KwTo $n$}{
        $r$ = sample a random walk of length $l$ originating from $v$ with
          transition probabilities determined by Equation~\ref{transition_probabilities}\\
        append $r$ to $R$
    }
}

{\bf return} word2vec($R$) 

\caption{{\bf The lid-n2v-rw algorithm}}	
\SetAlgoRefName{Al2}
\SetAlgoCaptionSeparator{'.'}
\end{algorithm}

It can be noticed that \verb|lid-n2v-rw| samples a proportionally higher number of random walks 
for high NC-LID nodes while keeping the computational budget (the total number of random walk steps per node)
approximately constant. The main idea is to increase the frequency of high NC-LID nodes in sampled
random walks in order to better preserve their close neighborhood in formed embeddings. Additionally,
the probability of the random walk leaving the natural community of the starting node
is lowered for high NC-LID nodes due to shorter random walks.

The second LID-elastic variant of node2vec, denoted by \verb|lid-n2v-rwpq|, extends \verb|lid-n2v-rw|
by adjusting $p$ and $q$ parameters controlling biases when sampling random walks. The pseudo-code for 
\verb|lid-n2v-rwpq| is given in Algorithm~3.
The original formula for determining transition probabilities in node2vec (Equation~\ref{transition_probabilities}) 
changes to
\begin{equation}
\label{transition_probabilities_2}
Pr(v \rightarrow x) = 
\begin{cases}
1/p(v, x)       & \mbox{if distance}(t, x) = 0 \\
1   & \mbox{if distance}(t, x) = 1\\
1/q(v, x) & \mbox{if distance}(t, x) = 2
\end{cases}
\end{equation}
where $p(v, x)$ and $q(v, x)$ are adjusted hyperparameters from their base values denoted by $p_{b}$ and
$q_{b}$, respectively (by default $p_{b} = q_{b} = 1$). The adjustments are made according to the 
following rules:
\begin{equation}
\label{adjustment_p}
p(v, x) = 
\begin{cases}
p_{b}                         & \mbox{if} \: v \: \mbox{is in the natural community of} \: x  \\
p_{b} + \mbox{NC-LID}(x)      & \mbox{otherwise}
\end{cases}
\end{equation}
\begin{equation}
\label{adjustment_q}
q(v, x) = 
\begin{cases}
q_{b}                         & \mbox{if} \: x \: \mbox{is in the natural community of} \: v  \\
q_{b} + \mbox{NC-LID}(v)      & \mbox{otherwise}
\end{cases}
\end{equation}


\begin{algorithm}[htb!]
\small
\SetAlgoLined
\DontPrintSemicolon
\SetKwInOut{Input}{input}
\SetKwInOut{Output}{output}
\Input{$G$ -- a graph}
\Output{$E$ -- an embedding of $G$} 

\BlankLine \BlankLine 
$R$ = empty list of sampled random walks \\
$V$ = set of nodes in $G$ \\
\ForEach{$v \in V$} {
    // determine the number of random walks that will be sampled for $v$ \\
    $n = \mbox{NRW}(v)$ (Equation~\ref{NRW})\\
    \BlankLine
    
    // determine the length of random walks sampled for $v$ \\
    $l = \mbox{LRW}(v)$ (Equation~\ref{LRW})\\
    
    \BlankLine
    // sample $n$ random walks \\
    \For{$i = 1$ \KwTo $n$}{
        $r$ = empty random walk starting at $v$\\
        \BlankLine
        
        // initialize current node $c$ to $v$ \\
        c = $v$
        \BlankLine
        
        // sample random walk of length $l$ \\
        \For{$j = 1$ \KwTo $l$}{            
            \ForEach{neighbor $x$ of $c$} {
                \If{$c$ is not in natural community of $x$} {
                    // this adjustment effectively affects only \\ 
                    // previous node in walk \\
                    adjust $p(c, x)$ according to Equation~\ref{adjustment_p}
                }
                \If{$x$ is not in natural community of $c$} {
                    adjust $q(c, x)$ according to Equation~\ref{adjustment_q}
                }
            }
            
            append $c$ to $r$ \\
            $c$ = sample node according to transition probabilities 
               given by Equation~\ref{transition_probabilities_2}\\
        }
        
        append $r$ to $R$
    }
}

{\bf return} word2vec($R$) 

\caption{{\bf The lid-n2v-rwpw algorithm}}	
\SetAlgoRefName{Al3}
\SetAlgoCaptionSeparator{'.'}
\end{algorithm}

The first rule given by Equation~\ref{adjustment_p} controls the probability of returning back from 
$v$ to $x$ if the random walk transitioned from $x$ to $v$ in the previous step. 
By increasing the base $p$ value if $v$ is not
in the natural community of $x$, \verb|lid-n2v-rwpq| lowers the probability of making a transition
between different natural communities. The increase is equal to $\mbox{NC-LID}(x)$ which implies that
the backtrack step is penalized more if $x$ has a more complex natural community.

The second rule (Equation~\ref{adjustment_q}) controls the probability of going to nodes that are more distant 
from the previously visited node in the random walk. The base $q$ value is increased for nodes not belonging to the natural
community of $v$ meaning that again \verb|lid-n2v-rwpq| penalizes transitioning between different
natural communities. The increase of $q_{b}$ is equal to $\mbox{NC-LID}(v)$ implying that
\verb|lid-n2v-rwpq| biases the random walk to stay within more complex natural communities.

\section{Experiments and Results}
\label{experiments_results}

Our experimental evaluation of the NC-LID measure and LID-elastic node2\-vec extensions is conducted on a corpus of real-world
graphs listed in Table~\ref{table_datasets}. The corpus contains 3 small graphs from the social network analysis 
literature (Zachary karate club, Les miserables and Florentine families) and 15 large-scale graphs that belong to the 
following categories of complex networks:
\begin{itemize}
 \item[--] Paper citation networks (CORAML, CORA, CITESEER, PUBMED, DB\-LP, cit-HepPh and cit-HepTh). Nodes in a paper 
 citation network correspond to scientific papers. Two papers $A$ and $B$ are connected 
 by a directed link $A \rightarrow B$ if $A$ cites $B$. 
 CORA, CITESEER, PUBMED and DBLP are citation networks named after bibliographic databases indexing research papers that were 
 used to extract respective networks. 
 CORA, DBLP and CITESEER primarily index computer science publications, while PUBMED stores metadata about biomedical 
 research papers. CORA\-ML is a subgraph of CORA induced by papers from the machine learning research field. 
 The previous citation networks are commonly used to evaluate graph neural networks~\cite{shchur2018pitfalls} 
 and graph embedding methods based on deep learning techniques~\cite{BojchevskiG18}.
 Additionally, we include two paper citation networks from the SNAP repository~\cite{snapnets} covering 
 research works in high energy physics (cit-HepPh and cit-HepTh). 
 
 \item[--] Scientific co-authorship networks (AstroPh, CondMat, GrQc and HepPh) from the SNAP repository~\cite{snapnets}.
 Those networks depict co-authorship relations among researchers. Two researchers, represented by two nodes in a co-authorship network,
 are connected by an undirected link if they published at least one research paper together. 
 Co-authorship networks are typically used to study the social structure and dynamics of 
 science~\cite{Savic2019Coauthorship,Savic2014Coauthorship}.
 
 \item[--] Co-purchase networks (AE photo and AE computers) that reflect similarities of e-commerce products. 
 Two products are joined by an undirected link if they are frequently bought together. AE photo and AE computers are co-purchase networks
 of Amazon photo and computer products, respectively. Those networks also frequently used in the evaluation of graph neural networks 
 and graph representational learning algorithms~\cite{shchur2018pitfalls,BojchevskiG18}. 
 
 \item[--] Social networks of social media users (BlogC and FB) available from the node2vec repository~\cite{nodetovec2016}. 
 BlogC is a large-scale social network describing social relationships of bloggers listed in the BlogCatalog website, while FB depicts
 friendships within a set of Facebook users.
\end{itemize}
In our experimental evaluation all directed graphs (paper citation networks) are converted to their 
undirected projections for two reasons: 
\begin{enumerate}
 \item the orientation of links is irrelevant for the notion of natural communities 
       and, consequently, for the definition of the NC-LID measure, and
 \item graph embedding methods based on random walks (e.g., node2vec) are usually applied on undirected projections 
       of directed graphs in order to capture neighborhoods containing both in-reachable and out-reachable nodes.
\end{enumerate}

\begin{table}[htb!]
\caption{Experimental datasets.}
\begin{center}
\begin{tabular}{lllllll}
\noalign{\smallskip}\hline \noalign{\smallskip}
Graph & $N$ & $L$ & $C$ & $F$ & $\bar{d}$ & $S$ \\
\noalign{\smallskip}\hline \noalign{\smallskip}
Zachary karate club & 34 & 78 & 1 & 1.00 & 4.59 & 2.00  \\
Les miserables & 77 & 254 & 1 & 1.00 & 6.60 & 1.89 \\
Florentine families & 15 & 20 & 1 & 1.00 & 2.67 & 0.62 \\
CORAML & 2995 & 8158 & 61 & 0.94 & 5.45 & 12.28 \\
CITESEER & 4230 & 5337 & 515 & 0.40 & 2.52 & 8.44 \\
AE photo & 7650 & 119081 & 136 & 0.98 & 31.13 & 10.42 \\
AE computers & 13752 & 245861 & 314 & 0.97 & 35.76 & 17.34 \\
PUBMED & 19717 & 44324 & 1 & 1.00 & 4.50 & 5.21 \\
CORA & 19793 & 63421 & 364 & 0.95 & 6.41 & 7.87 \\
DBLP & 17716 & 52867 & 589 & 0.91 & 5.97 & 9.43 \\
AstroPh  &  18772  &  198050  &  290  &  0.95  &  21.10  &  3.85 \\
CondMat  &  23133  &  93439  &  567  &  0.92  &  8.08  &  5.73 \\
GrQc  &  5242  &  14484  &  355  &  0.79  &  5.53  &  3.83 \\
HepPh  &  12008  &  118489  &  278  &  0.93  &  19.74  &  5.02 \\
cit-HepPh  &  34546  &  420877  &  61  &  0.99  &  24.37  &  5.22 \\
cit-HepTh  &  27770  &  352285  &  143  &  0.99  &  25.37  &  17.19 \\
BlogC  &  10312  &  333983  &  1  &  1.00  &  64.78  &  9.82 \\
FB  &  4039  &  88234  &  1  &  1.00  &  43.69  &  4.52 \\
\noalign{\smallskip}\hline \noalign{\smallskip}
\end{tabular}
\label{table_datasets}
\end{center}
\end{table}

For each graph, Table~\ref{table_datasets} shows 
the number of nodes ($N$), 
the number of links ($L$), 
the number of connected components ($C$),
the fraction of nodes in the largest connected component ($F$), 
the average degree ($\bar{d}$) and
the skewness of the degree distribution ($S$).
It can be seen that all graphs are sparse ($\bar{d} \ll N - 1$), which is a typical characteristic of real-world networks.
Second, all graphs, except CITESEER, are either connected graphs ($C = 1$) 
or have a giant connected component ($F > 0.75$).
The degree distributions of 15 large-scale graphs have a considerably high positive skewness 
implying that they are heavy-tailed. This means that the corresponding graphs contain 
so-called hubs -- nodes having a large number of neighbors that is significantly higher 
than the average degree.

\subsection{Natural Communities and NC-LID}

\begin{figure}[htb!]
\begin{center}
\includegraphics[width=0.95\textwidth]{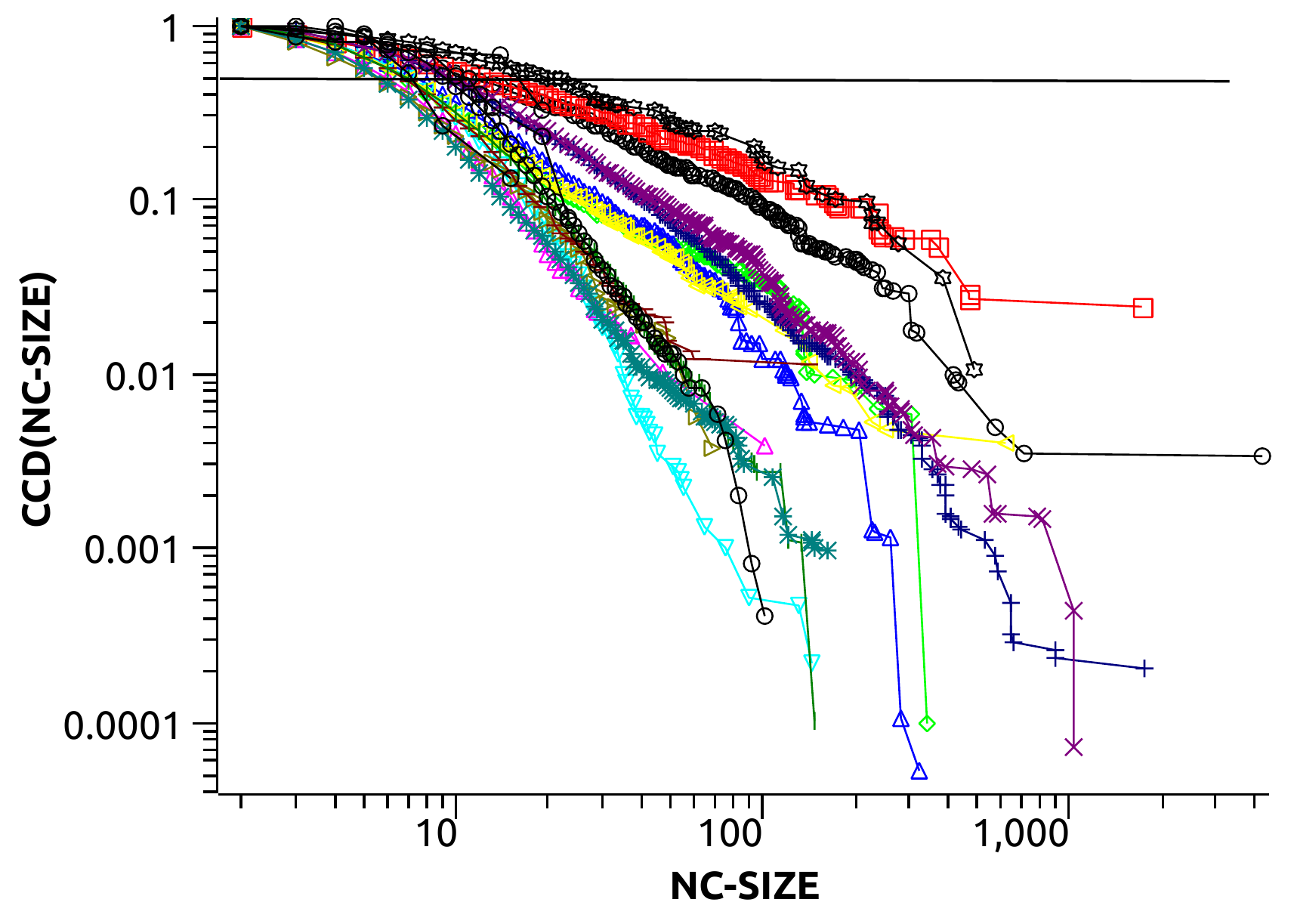}
\end{center}
\caption{The complementary cumulative distribution of NC-SIZE for all graphs from 
our corpus. The solid horizontal line represents probability 0.5.}
\label{fig_ncdistr}
\end{figure}

The NC-LID measure is defined considering natural communities in a graph.
The most basic characteristic of a natural community is its size, i.e. the number 
of nodes it contains (denoted by NC-SIZE). Since nodes in real-world graphs exhibit extremely 
diverse connectivity characterized by heavy-tailed degree distributions, it
can be also expected that natural communities also considerably vary in 
their size. Figure~\ref{fig_ncdistr} shows the complementary cumulative 
distribution of NC-SIZE, denoted by CCD$(s)$, for all 
graphs from our experimental corpus on a log-log plot. 
CCD($s$) is the probability of 
observing a natural community that contains $s$ or more nodes.
It can be seen that empirically observed CCDs have very long tails. 
This implies that majority of nodes have relatively small natural communities, 
but there are also nodes having exceptionally large natural communities encompassing 
hundreds or even thousands of nodes. For example, 53.17\% of cit-HepPh nodes have natural 
communities with less than 10 nodes, while the largest natural community in this graph contains 
1744 nodes. Considering all graphs from our experimental corpus, natural 
communities typically contain between 6 and 21 nodes 
(CCD$(s) < 0.5$ for $s \in [6, 21]$, see Figure~\ref{fig_ncdistr}). On the other hand,
4 graphs have natural communities with more than 1000 nodes (AE computers, AE photo, cit-HepTh, 
cit-HepPh). The largest natural community in all other graphs, except the 3 smallest ones 
and CITESEER, encompasses more than 100 nodes. 

\begin{figure}[htb!]
\begin{center}
\includegraphics[width=0.95\textwidth]{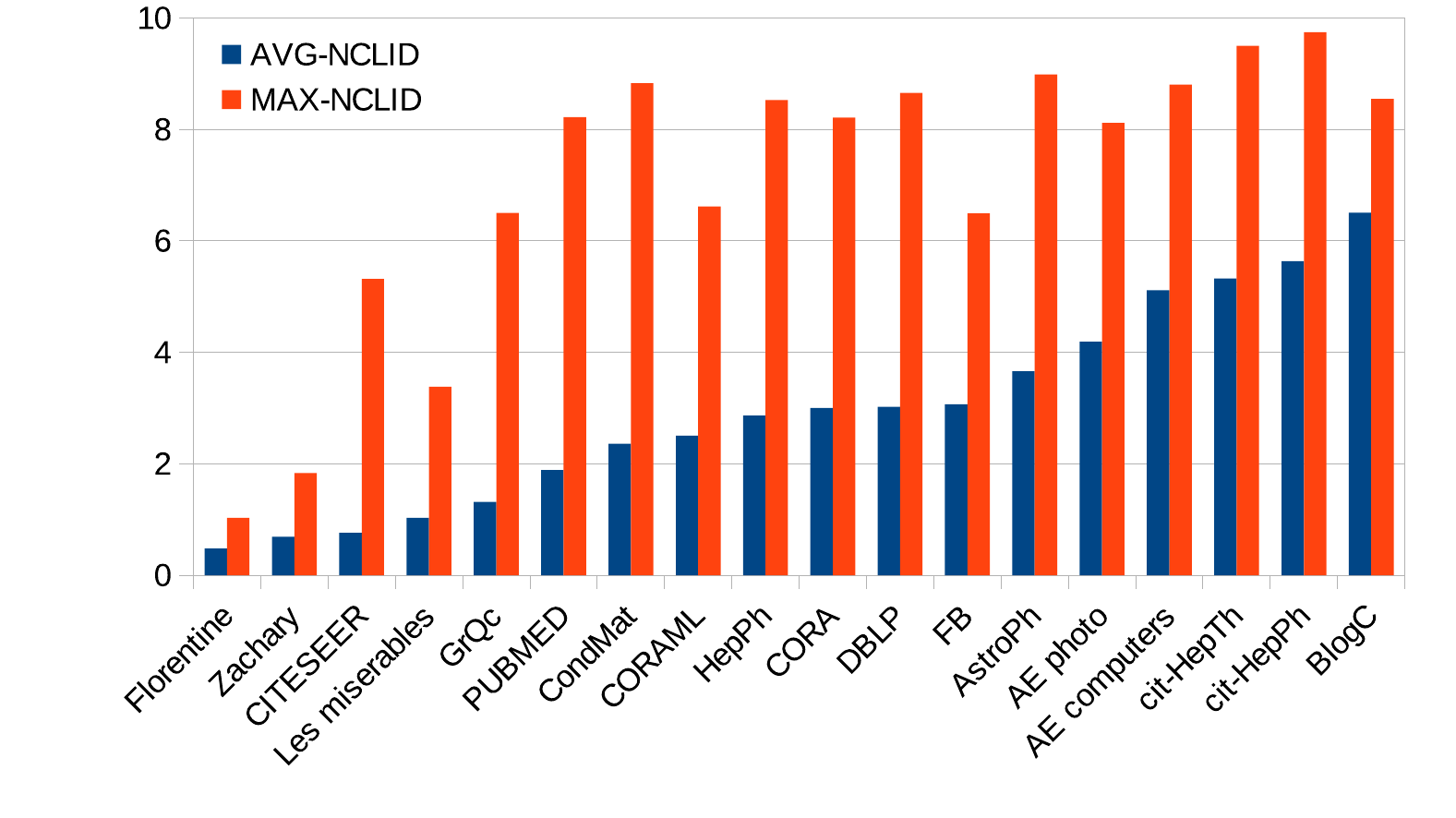}
\end{center}
\caption{The average and the maximal NC-LID for graphs from our experimental corpus.}
\label{fig_nclid}
\end{figure}

The average and maximal NC-LID in examined graphs are shown in Figure~\ref{fig_nclid}
sorted by the average NC-LID from the graph with the most compact natural communities 
to the graph with the most complex natural communities on average. 
The minimal NC-LID in all graphs is equal to 0, which is the lowest possible NC-LID. 
The lowest possible NC-LID corresponds to nodes whose local communities are strongly compact or convex
in the sense that distances from a node to all nodes from its natural community are strictly smaller 
than distances from the node to nodes not belonging to its natural community 
(i.e., the shortest-path distance measure perfectly separates nodes from the natural community 
from outside nodes). 
It should be noticed that the average/maximal NC-LID does not steadily increase with
the number of nodes nor with the density (average degree). In other words, smaller (resp., sparser)
graphs may have more complex natural communities than larger (resp., denser) graphs. 
The social network of Florentine families has the lowest
average NC-LID equal to 0.48. This NC-LID level means that approximately 38\% of nodes within
the shortest-path radius of the natural community of a randomly selected node do not belong to
its natural community. BlogC has the most complex natural communities with the largest average 
NC-LID equal to 6.51. This NC-LID value corresponds to situations in which approximately 
0.14\% of nodes within the shortest-path radius of a natural community belong to the 
natural community. 

\begin{figure}[htb!]
\begin{center}
\includegraphics[width=0.8\textwidth]{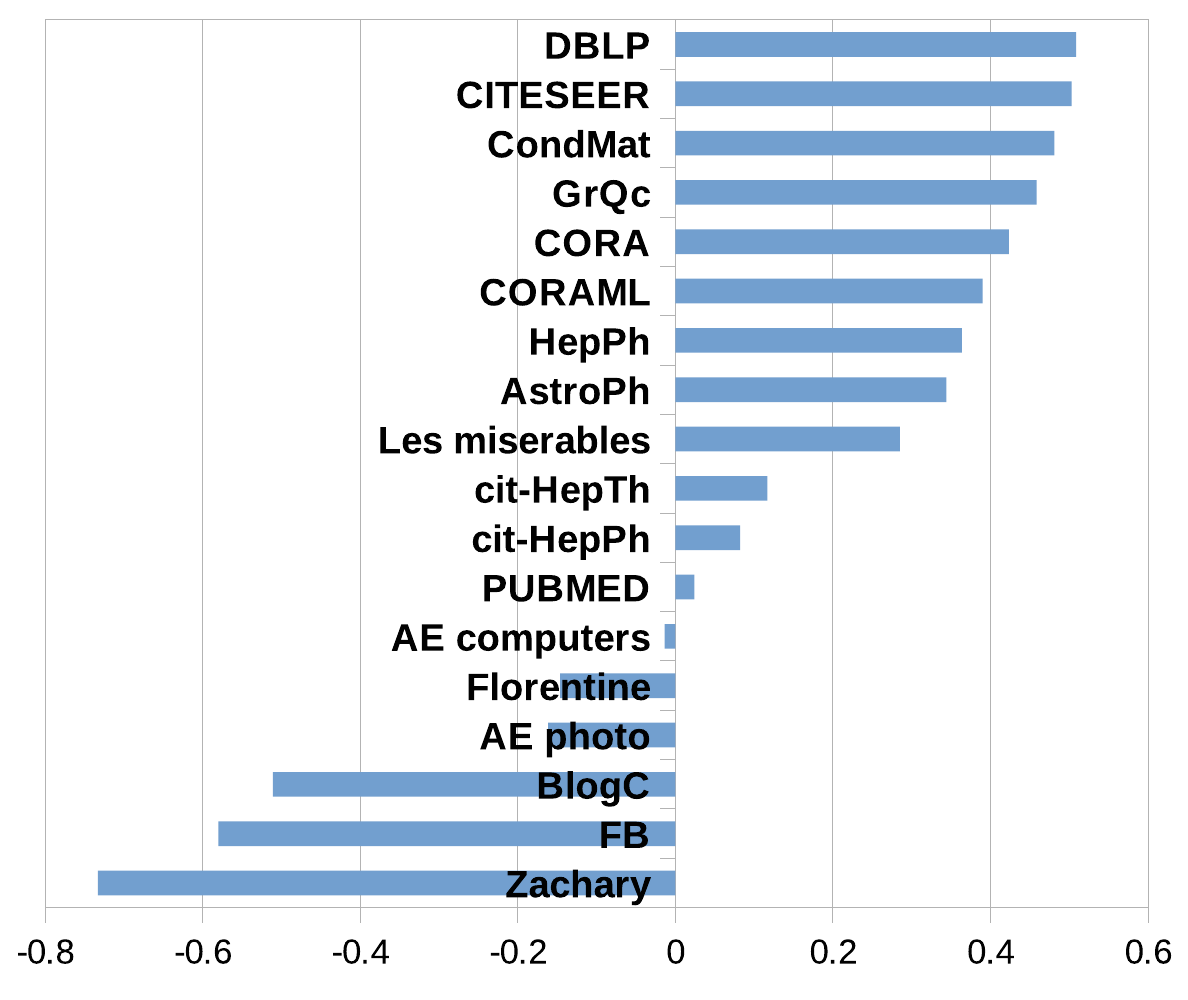}
\end{center}
\caption{Spearman correlation between NC-LID and NC-SIZE.}
\label{fig3_nclid_ncs}
\end{figure}

It is important to emphasize that NC-LID does not necessarily increase with NC-SIZE, 
i.e. smaller natural communities may be more complex than larger natural communities. 
Figure~\ref{fig3_nclid_ncs} shows the values of the Spearman correlation coefficient 
between NC-LID and NC-SIZE. It can be seen that for 8 (out of 18) graphs there is a 
positive correlation higher than 0.3. In those graphs larger natural communities tend to be more 
complex than smaller natural communities. The opposite tendency is present for 3 graphs 
(BlogC, FB and Zachary) in which smaller natural communities tend to be more complex than 
larger natural communities (a significant negative correlation lower than \num{-0.3}). 
For AE photo and PUBMED, Spearman correlation is close to 0 implying that 
larger natural communities do not tend to be neither more nor less complex 
than smaller natural communities.  

To better understand the structural characteristics of nodes with complex natural communities we
examine correlations between NC-LID and node centrality metrics quantifying structural importance of nodes
in complex networks~\cite{Savic2019Fundametal}. Node centrality metrics can be divided into 
two categories: 
\begin{enumerate}
\item local metrics taking into account only links emanating from a node, and 
\item global metrics reflecting the importance of the node considering the complete structure of the network. 
\end{enumerate}
The degree of a node (denoted by DEG) is the most basic local centrality metric. The main advantage of this metric 
is that it is easy and fast to compute. Nodes having high degree are also called hubs and they are especially 
important for the overall connectivity in scale-free networks and networks with heavy-tailed degree distributions, 
in general. 

In our experimental evaluation we consider the following global centrality metrics: 
\begin{itemize}
\item[--] the shell or core index (denoted by CORE), 
\item[--] eigenvector centrality (denoted by EVC),
\item[--] closeness centrality (denoted by CLO), and 
\item[--] betweenness centrality (denoted by BET).
\end{itemize}

The core index is a metric related to the $k$-core decomposition of networks. 
The $k$-core is of a graph is its subgraph obtained by repeatedly removing nodes whose degree is smaller than $k$.
The core index of a node is equal to $k$ if the node belongs to the $k$-core, but not to the $(k + 1)$-core. 
Hubs predominantly connected to other hubs tend to have high values of this metric. 

The eigenvector centrality metric is based on the principle that important nodes tend to be connected to other important nodes. 
This principle when expressed as a recurrence relation yields the eigenvector of the adjacency matrix as 
the vector expressing structural importance of nodes. 

The intuition behind closeness centrality is that important nodes tend to be located in proximity to many other nodes. 
Formally, the closeness centrality of a node is inversely proportional to the total shortest-path distance between 
the node and all other nodes in the network. 

The betweenness centrality of the node is the extent to which the node tends to be located 
on the shortest path between any two arbitrary nodes. Nodes with high betweenness tend to connect different 
cohesive node groups or clusters in the network, whereas nodes with high closeness tend 
to be the most central nodes in the most central clusters.  

The values of Spearman correlation between NC-LID and node centrality metrics
are shown in Table~\ref{table_NCLID_centrality}. 
Except in a few cases (EVC for Zachary; DEG and CORE for FB), 
there are positive Spearman correlations between NC-LID and node centrality metrics. 
It can be observed that DEG tends to have moderate Spearman correlations 
with NC-LID: for 7 graphs the correlation coefficient takes values between 0.4 and 0.6, 
for 9 graphs between 0.2 and 0.4 and only for 2 graphs there are weak correlations lower than 0.2. 
Moderate correlations lower than 0.6 are also present for BET. CORE, EVC and CLO tend to exhibit 
considerably higher correlations with NC-LID compared to DEG and BET. 
This result implies that nodes having complex local communities 
tend to be more globally than locally important. However, their global importance is not 
determined by their brokerage role in connecting different node clusters (high BET), 
but by their central positions (high CLO) in the most important clusters (high EVC) 
that are located in the core of the network (high CORE). 
For all datasets except one (Zachary, the smallest graph from our experimental corpus), 
CLO shows the strongest correlations with NC-LID: 
the correlation between NC-LID and CLO is very strong (higher than 0.8) 
for two graphs (CITESEER and GrQc), strong (between 0.6 and 0.8) for 7 graphs, 
moderate (between 0.4 and 0.6) for 6 graphs and only for 3 graphs 
correlations can be considered as low to moderate 
(values lower than 0.4 but higher than 0.2). 
This result indicates that high NC-LID nodes tend to be the most central nodes 
that have many long-range links, i.e., links whose removal drastically increases 
distances between nodes. Such long-range links are actually links connecting nodes 
belonging to different node clusters implying that high NC-LID nodes tend to be 
located in the most central node clusters 
(as also indicated by high correlations between NC-LID and CORE/EVC). 
Additionally, high NC-LID nodes tend to be well connected to both nodes 
from their own node clusters and nodes belonging to different node clusters which 
can explain high complexity or ``concavity'' of their natural communities 
since long-range links lead outside of the natural community of a node.

\begin{table}[htb!]
\caption{Spearman correlations between NC-LID and various centrality metrics quantifying structural importance of nodes in complex networks.}
\begin{center}
\begin{tabular}{llllll}
\noalign{\smallskip}\hline \noalign{\smallskip}
Graph & DEG & CORE & EVC & CLO & BET \\
\noalign{\smallskip}\hline \noalign{\smallskip}
Zachary & 0.113 & 0.137 & -0.175 & 0.231 & 0.248  \\
Florentine & 0.200 & 0.429 & 0.435 & 0.535 & 0.240 \\
Les miserables & 0.302 & 0.269 & 0.441 & 0.444 & 0.349 \\
CORAML & 0.379 & 0.462 & 0.609 & 0.667 & 0.340 \\
CITESEER & 0.412 & 0.519 & 0.807 & 0.819 & 0.459 \\
PUBMED & 0.414 & 0.498 & 0.537 & 0.541 & 0.366 \\
CORA & 0.434 & 0.518 & 0.569 & 0.681 & 0.339 \\
DBLP & 0.536 & 0.615 & 0.712 & 0.742 & 0.397 \\
AE photo & 0.356 & 0.396 & 0.543 & 0.601 & 0.240 \\
AE computers & 0.441 & 0.478 & 0.564 & 0.588 & 0.283 \\ 
BlogC & 0.239 & 0.250 & 0.263 & 0.267 & 0.168 \\
AstroPh & 0.437 & 0.434 & 0.625 & 0.658 & 0.372 \\
CondMat & 0.398 & 0.392 & 0.634 & 0.689 & 0.355 \\
GrQc & 0.350 & 0.343 & 0.708 & 0.802 & 0.410 \\
HepPh & 0.367 & 0.376 & 0.652 & 0.691 & 0.338 \\
cit-HepPh & 0.360 & 0.407 & 0.446 & 0.465 & 0.226 \\
cit-HepTh & 0.472 & 0.508 & 0.563 & 0.578 & 0.266 \\
FB & -0.004 & -0.011 & 0.092 & 0.231 & 0.122 \\
\noalign{\smallskip}\hline \noalign{\smallskip}
\end{tabular}
\label{table_NCLID_centrality}
\end{center}
\end{table}

\subsection{Node2vec Evaluation}

In our experimental evaluation, node2vec was tuned by finding values of 
its hyper-parameters $p$ (return-back parameter) 
and $q$ (in-out parameter) that give embeddings maximizing the $F_{1}$ score
in five different graph embedding dimensions (10, 25, 50, 100 and 200).  
As suggested by the authors of node2vec~\cite{nodetovec2016}, 
for $p$ and $q$ we considered values in \{0.25, 0.50, 1, 2, 4\}, while the number of 
random walks per node and the length of each random walk were fixed to 10 and 80, 
respectively.

The results of node2vec tuning are presented in Table~\ref{table_n2v_bestembs}. 
The table shows maximal $F_{1}$ scores for graphs from our experimental corpus sorted 
from the largest to the lowest score, the dimension in which the maximal score 
is achieved (Dim.), the corresponding values of $p$ and $q$, 
and graph reconstruction evaluation metrics ($P$ -- link precision and $R$ -- link recall). 
It can be observed that the maximal $F_{1}$ score is in the range $[0.24, 0.96]$. 
For the majority of graphs the maximal $F_{1}$ is larger than 0.5 implying that node2vec 
graph embeddings preserve the structure of input graphs to a very good extent. 
It can also be seen that link precision and recall tend to exhibit high correlations with $F_{1}$: 
a larger $F_{1}$ score in the vast majority of cases implies larger link precision and 
recall (the values of Person's correlation coefficient between both link precision and recall 
on the one side and $F_{1}$ on the other side is equal to 0.98).

\begin{table}[htb!]
\caption{Characteristics of the best node2vec embeddings (sorted from the highest to the lowest $F_{1}$ score).}
\begin{center}
\begin{tabular}{lllllll}
\noalign{\smallskip}\hline \noalign{\smallskip}
Graph & Dim. & $p$ & $q$ & $P$ & $R$ & $F_{1}$ \\
\noalign{\smallskip}\hline \noalign{\smallskip}
Florentine families & 100 & 0.25 & 4 & 0.97 & 0.96 & 0.96     \\
Les miserables & 100 & 0.25 & 4 & 0.79 & 0.83 & 0.81 \\
Zachary karate club & 100 & 0.25 & 4 & 0.78 & 0.78 & 0.78 \\
AstroPh & 50 & 2 & 0.25 & 0.68 & 0.74 & 0.71 \\
HepPh & 25 & 0.5 & 0.25 & 0.68 & 0.73 & 0.71 \\
CondMat & 25 & 0.5 & 4 & 0.69 & 0.62 & 0.65 \\
CORAML & 25 & 0.5 & 0.25 & 0.63 & 0.67 & 0.65 \\
FB & 25 & 0.25 & 2 & 0.64 & 0.63 & 0.64 \\
CORA & 25 & 4 & 0.25 & 0.58 & 0.56 & 0.57 \\
cit-HepTh & 100 & 2 & 0.25 & 0.56 & 0.59 & 0.57 \\
GrQc & 10 & 0.25 & 2 & 0.61 & 0.53 & 0.56 \\
cit-HepPh & 100 & 4 & 0.25 & 0.54 & 0.51 & 0.53 \\
AE photo & 50 & 0.5 & 0.5 & 0.51 & 0.48 & 0.5 \\ 
AE computers & 50 & 4 & 0.25 & 0.49 & 0.42 & 0.45 \\
DBLP & 25 & 0.5 & 1 & 0.44 & 0.37 & 0.4 \\
PUBMED & 50 & 4 & 0.25 & 0.32 & 0.52 & 0.39 \\ 
BlogC & 50 & 0.25 & 0.25 & 0.28 & 0.22 & 0.25 \\ 
CITESEER & 10 & 0.5 & 0.25 & 0.23 & 0.24 & 0.24 \\ 
\noalign{\smallskip}\hline \noalign{\smallskip}
\end{tabular}
\label{table_n2v_bestembs}
\end{center}
\end{table}

\begin{figure}[htb!]
\begin{center}
\includegraphics[width=0.85\textwidth]{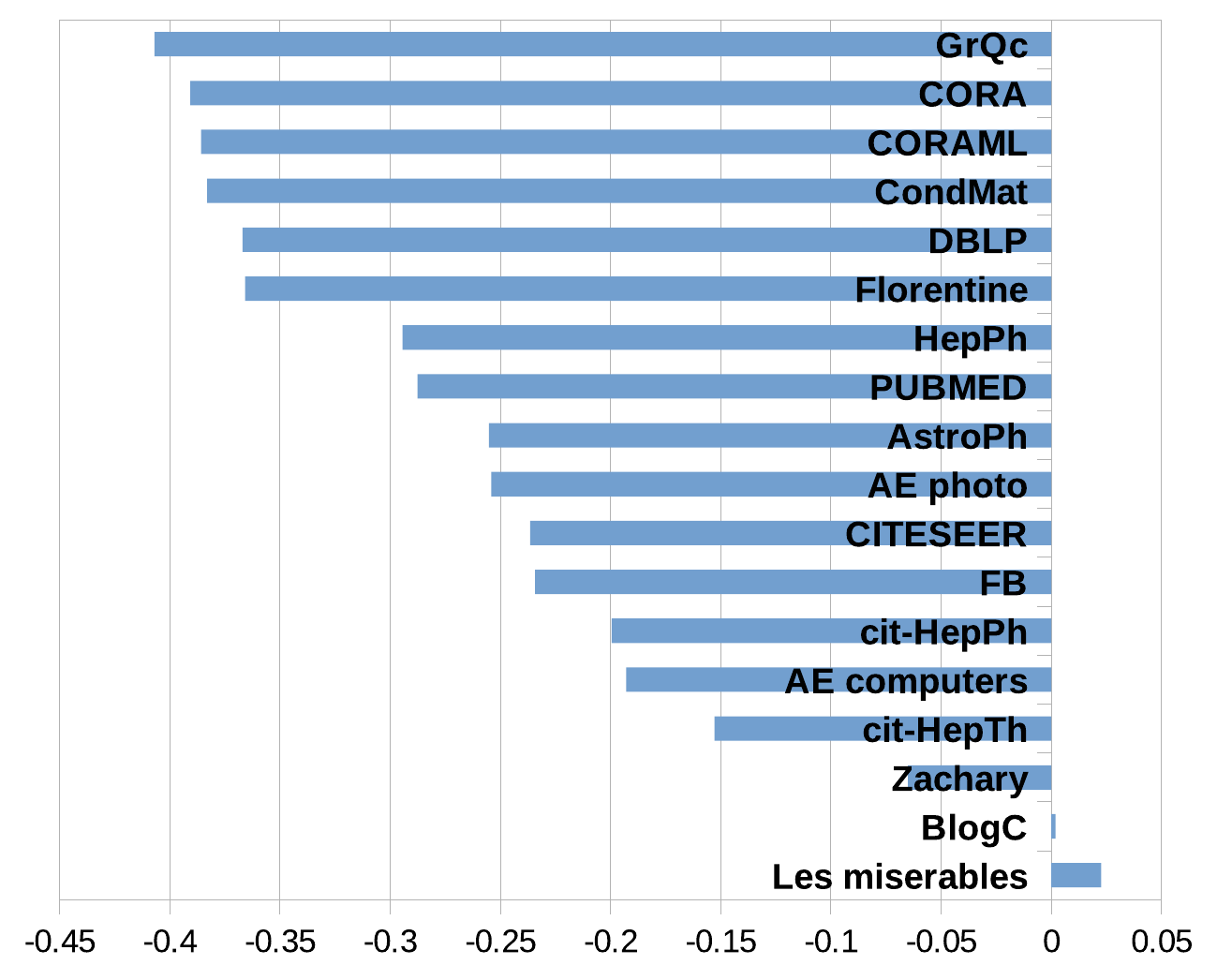}
\end{center}
\caption{The Spearman correlation between NC-LID and $F_{1}$ scores in the
best node2vec embeddings.}
\label{fig_SC_NCLID_F1}
\end{figure}

High NC-LID nodes have more complex natural communities than low NC-LID nodes. 
It is reasonable to expect that high NC-LID will have higher reconstruction errors 
in graph embeddings due to more complex natural communities. To check 
this assumption we analyze Spearman correlations between NC-LID of nodes 
and their $F_{1}$ scores in the best node2vec embeddings obtained after 
hyper-parameter tuning. The values of the Spearman correlation coefficient between 
NC-LID and $F_{1}$ for all graphs from our experimental corpus are shown in Figure~\ref{fig_SC_NCLID_F1}. 
It can be seen that for 15 (out of 18) graphs there are significant 
negative Spearman correlations ranging between 
\num{-0.2} and \num{-0.4}. Consequently, it can be concluded that NC-LID is able to fairly 
accurately identify nodes that have high link reconstruction errors 
(please recall that lower $F_{1}$ scores imply higher link reconstruction errors).

To further examine how NC-LID is related to the precision of links reconstructed from embeddings we divide nodes into two groups: 
\begin{enumerate}
\item nodes having NC-LID above the average NC-LID (high NC-LID nodes, group $H$), and 
\item nodes having NC-LID below the average NC-LID (low NC-LID nodes, group $L$).
\end{enumerate}
Then, we compare $F_{1}$ scores of those two groups of nodes using the Mann-Whitney U (MWU) test~\cite{MWU1947}. 
The MWU test is a test of stochastic equality checking the null hypothesis stating that 
$F_{1}$ scores in $H$ do not tend to be neither greater nor smaller than $F_{1}$ scores in $L$. 
Additionally, we compute two probabilities of superiority reflecting 
tendencies of stochastic inequalities: 
\begin{enumerate}
\item PS($H$) -- the probability that a randomly selected node from $H$ has a higher $F_{1}$ than a randomly selected node form $L$,
\item PS($L$) - the probability that a randomly selected node from $L$ has a higher $F_{1}$ than a randomly selected node from $H$ 
($\mbox{PS}(L) = 1 - \mbox{PS}(H) - P_{e}$, where $P_{e}$ is the probability of observing equal $F_{1}$ scores in $H$ and $L$).
\end{enumerate}
The results of statistical testing and accompanying probabilities of superiority are presented in Table~\ref{table_mwu}.  
$F_{1}(H)$ and $F_{1}(L)$ in Table~\ref{table_mwu} denote the average $F_{1}$ score for nodes in $H$ and $L$, respectively. 
$U$ is the value of the MWU test statistic and the null hypothesis of the test is accepted if the $p$-value of $U$ is higher 
than 0.05, which is also indicated by the ACC column in Table~\ref{table_mwu} (``yes'' means that the null hypothesis is accepted). 
From the obtained results we can observe that the null hypothesis is accepted for the three smallest graphs: Zachary karate club, 
Les miserables and Florentine families. For those three graphs differences in $F_{1}$ scores of high and low NC-LID nodes 
are not statistically significant. For large graphs we have that $F_{1}$
scores of high NC-LID nodes tend to be significantly lower than $F_{1}$ scores of low NC-LID nodes: 
$F_{1}(H) < F_{1}(L)$ and $\mbox{PS}(H) \ll \mbox{PS}(L)$.

\begin{table}[htb!]
\caption{Comparison of $F_{1}$ scores of high NC-LID nodes ($H$) and
low NC-LID nodes ($L$) using the Mann-Whitney U test.}
\begin{center}
\begin{tabular}{llllllll}
\noalign{\smallskip}\hline \noalign{\smallskip}
Graph & $F_{1}(H)$ & $F_{1}(L)$ & $U$ & $p$ & ACC & PS($H$) & PS($L$) \\
\noalign{\smallskip}\hline \noalign{\smallskip}
Zachary & 0.70 & 0.71 & 132 & 0.44 & yes & 0.44 & 0.48  \\
Les miserables & 0.76 & 0.76 & 734 & 0.50 & yes & 0.47 & 0.47 \\
Florentine & 0.93 & 0.98 & 19 & 0.10 & yes & 0.07 & 0.39 \\
CORAML & 0.44 & 0.62 & 699380 & $< 0.01$ & no & 0.29 & 0.67 \\
CITESEER & 0.10 & 0.25 & 1707420 & $< 0.01$ & no & 0.19 & 0.31 \\
AE photo & 0.32 & 0.43 & 5239408 & $< 0.01$ & no & 0.36 & 0.64 \\
AE computers & 0.29 & 0.38 & 17900546 & $< 0.01$ & no & 0.38 & 0.61 \\
PUBMED & 0.19 & 0.32 & 31448278 & $< 0.01$ & no & 0.28 & 0.59 \\
CORA & 0.36 & 0.54 & 29695497 & $< 0.01$ & no & 0.28 & 0.68 \\
DBLP & 0.20 & 0.42 & 26684749 & $< 0.01$ & no & 0.25 & 0.57 \\
BlogC & 0.13 & 0.14 & 10915606 & $< 0.01$ & no & 0.45 & 0.49 \\
AstroPh & 0.55 & 0.65 & 32743550 & $< 0.01$ & no & 0.37 & 0.62 \\
CondMat & 0.45 & 0.62 & 42561679 & $< 0.01$ & no & 0.31 & 0.66 \\
GrQc & 0.35 & 0.57 & 2051967 & $< 0.01$ & no & 0.28 & 0.64 \\
HepPh & 0.53 & 0.66 & 12720029 & $< 0.01$ & no & 0.35 & 0.63 \\
cit-HepPh & 0.35 & 0.43 & 114704661 & $< 0.01$ & no & 0.39 & 0.60 \\
cit-HepTh & 0.41 & 0.47 & 79349590 & $< 0.01$ & no & 0.43 & 0.57 \\
FB & 0.51 & 0.59 & 1634998 & $< 0.01$ & no & 0.41 & 0.59 \\
\noalign{\smallskip}\hline \noalign{\smallskip}
\end{tabular}
\label{table_mwu}
\end{center}
\end{table}

The next question we empirically address is whether NC-LID is a better indicator of nodes 
with low $F_{1}$ scores in node2vec embeddings than node centrality metrics.
We have computed Spearman correlations (denoted by $\rho$) between node centrality metrics and NC-LID one one side 
and $F_{1}$ scores in the best node2vec embeddings on the other side for all graphs from our 
experimental corpus in order to determine which node metric has the strongest ability to 
point to nodes high link reconstruction errors. 
The obtained correlations are given in Table~\ref{table_general_correlations}. 
The first thing that should be noticed is that only NC-LID exhibits  
negative correlations with $F_{1}$, whereas for node centrality metrics it can be 
observed both negative and positive correlations. 
For example, DEG exhibits a positive correlation with $F_{1}$ 
on Blog-catalog ($\rho$ = 0.306) and a negative correlation on DBLP ($\rho$ = \num{-0.398}).  
When a metric exhibits both negative and positive correlations with $F_{1}$ 
then it cannot be solely exploited to improve a graph embedding algorithm. 
In cases of strong positive correlations, high values of the metric indicate 
``good'' nodes (nodes with low graph reconstruction errors), while in cases of  
strong negative correlations high values point out to ``bad'' nodes 
(nodes with high graph reconstruction errors). 

\begin{table}[htb!]
\caption{Spearman correlations between $F_{1}$ scores in the best node2vec embeddings and various
metrics for nodes in complex networks including NC-LID.}
\begin{center}
\begin{tabular}{lllllll}
\noalign{\smallskip}\hline \noalign{\smallskip}
Graph & NC-LID & DEG & CORE & EVC & CLO & BET \\
\noalign{\smallskip}\hline \noalign{\smallskip}
Zachary & \num{-0.065} & 0.193 & 0.173 & \num{-0.125} & \num{-0.03} & 0.071   \\
Florentine & \num{-0.367} & \num{-0.352} & \num{-0.423} & \num{-0.466} & \num{-0.25} & \num{-0.158} \\ 
Les miserables & \num{-0.045} & 0.308 & 0.396 & 0.118 & \num{-0.121} & 0.002 \\
CORAML & \num{-0.386} & \num{-0.112} & \num{-0.178} & \num{-0.318} & \num{-0.421} & \num{-0.129} \\
CITESEER & \num{-0.236} & \num{-0.161} & \num{-0.108} & \num{-0.307} & \num{-0.317} & \num{-0.215} \\
PUBMED & \num{-0.288} & \num{-0.252} & \num{-0.346} & \num{-0.165} & \num{-0.248} & \num{-0.257} \\
CORA & \num{-0.391} & \num{-0.215} & \num{-0.255} & \num{-0.271} & \num{-0.384} & \num{-0.251} \\
DBLP & \num{-0.367} & \num{-0.398} & \num{-0.438} & \num{-0.31} & \num{-0.362} & \num{-0.392} \\
AE photo & \num{-0.264} & 0.088 & 0.053 & \num{-0.122} & \num{-0.196} & \num{-0.026} \\
AE computers & \num{-0.204} & 0.026 & \num{-0.014} & \num{-0.15} & \num{-0.208} & \num{-0.021} \\
BlogC & 0.002 & 0.306 & 0.305 & 0.286 & 0.269 & 0.304 \\
AstroPh & \num{-0.255} & 0.069 & 0.114 & \num{-0.096} & \num{-0.132} & \num{-0.16} \\
CondMat & \num{-0.383} & \num{-0.109} & \num{-0.015} & \num{-0.219} & \num{-0.305} & \num{-0.377} \\
GrQc & \num{-0.407} & \num{-0.046} & 0.036 & \num{-0.269} & \num{-0.34} & \num{-0.358} \\
HepPh & \num{-0.294} & 0.2 & 0.222 & \num{-0.088} & \num{-0.109} & \num{-0.161} \\
cit-HepPh & \num{-0.199} & 0.041 & 0.028 & \num{-0.029} & \num{-0.098} & \num{-0.054} \\
cit-HepTh & \num{-0.153} & 0.087 & 0.072 & 0.015 & \num{-0.026} & 0.009 \\
FB & \num{-0.234} & 0.452 & 0.449 & 0.157 & 0.06 & 0.166 \\
\noalign{\smallskip}\hline \noalign{\smallskip}
\end{tabular}
\label{table_general_correlations}
\end{center}
\end{table}

Node metric $M_{1}$ better points to weak parts of a node2vec embedding (nodes with low $F_{1}$ scores) 
obtained from a graph $G$ compared to node metric $M_{2}$ if $M_{1}$ achieves stronger 
correlations with $F_{1}$ (by absolute value) than $M_{2}$ with $F_{1}$. 
In such cases we say that $M_{1}$ wins over $M_{2}$ on $G$.  
From the results shown in Table 4 it can be seen that:
\begin{enumerate}
\item When NC-LID wins over some other metric, it is always a winning situation with a significantly stronger negative correlation.
\item NC-LID achieves 13 wins over DEG. DEG wins only on 5 datasets: on 1 dataset with a significant positive correlation and 
on 4 datasets with significant negative correlations.  
\item NC-LID wins 11 times over CORE. CORE is better than NC-LID on 7 datasets (including the three smallest graphs), 
3 times with significant negative and 4 times with significant positive correlations.
\item NC-LID is better than EVC on 13 graphs. EVC wins 3 times with significant negative correlations and 2 times with significant positive correlation.
\item NC-LID has 12 wins over CLO. CLO is better on 6 datasets: 5 times with negative correlations and once with a significant positive correlation (on Blog-catalog which is a large graph).
\item NC-LID is better than BET 15 times. BET has only 3 wins over NC-LID (two times with positive and once with a negative correlation).
\end{enumerate}

According to the number of wins, it can be concluded that NC-LID is a much better indicator of nodes with low $F_{1}$ scores in node2vec embeddings than node centrality metrics. Additionally, NC-LID can be computed more quickly than global centrality metrics. Finally, node centrality metrics in some cases are able to point to weak parts and in some other cases to good parts of node2vec embeddings. Therefore, they cannot be directly utilized to adjust node2vec parameters per node (the number and length of random walks) or pair of nodes (parameters controlling random walk biases) in contrast to NC-LID which always indicates bad parts of node2vec embeddings.

\subsection{LID-elastic node2vec Evaluation}
\label{lid_node2vec_eval}

Our LID-elastic node2vec extensions are based on the premise that high NC-LID nodes
have higher link reconstruction errors than low NC-LID nodes due to more complex
natural communities. The quality of graph embeddings can be assessed
by comparing original graphs to graphs reconstructed from embeddings. Let $G$ denote
an arbitrary graph with $L$ links and let $E$ be an embedding constructed
from $G$ using some graph embedding algorithm.
Please recall that $E$ is actually a list of real-valued vectors of the same size
(embedding dimension), one vector per node.
The graph reconstructed from $E$ has the same number of links as $G$.
The links in the reconstructed graph are formed by joining
the $L$ closest vector pairs from $E$ by Euclidean distance. 
The below defined link reconstruction error metrics can be used to
assess the quality of $E$ according to the principle that nodes close in 
$G$ should also be close in $E$.

\begin{definition}[{\it Link precision}]
The link precision for node $n$, denoted by $P(n)$, is equal to the number of correctly reconstructed
links incident to $n$ divided by the total number of links incident to $n$ in the reconstructed graph.
\label{def1}
\end{definition}

\begin{definition}[{\it Link recall}]
The link recall for node $n$, denoted by $R(n)$, is the number of correctly reconstructed links 
incident to $n$ divided by the total number of links incident to $n$ in the original graph. 
\label{def2}
\end{definition}

\begin{definition}[$F_{1}$ {\it score}]
The $F_{1}$ score for node $n$ aggregates link precision and recall into a single measure
by taking their harmonic mean, i.e.
\begin{equation*}
F_{1}(n) = \frac{2 \cdot P(n) \cdot R(n)}{P(n) + R(n)}.
\end{equation*}
\label{ref3}
\end{definition}
Higher values of $P(n)$, $R(n)$ and $F_{1}(n)$ imply lower link reconstruction errors for $n$.
The link reconstruction errors at the level of the whole graph can be obtained by 
taking averages across all nodes.

The obtained results of the evaluation of LID-elastic node2vec variants are summarized in Table~\ref{table_lid_n2v}.
The table shows the best $F_{1}$ scores of node2vec (n2v), the dimensions in which they are achieved (Dim.), 
the best $F_{1}$ scores of LID-elastic node2vec variants and corresponding embedding dimensions. 
The column ``Best'' indicates the best graph embedding algorithm achieving the maximal $F_{1}$ score 
(i.e., the algorithm that better preserves the graph structure compared to the others) where \texttt{rw} denotes 
the first LID-elastic node2vec extension (\texttt{lid-n2v-rw}) and \texttt{rwpq} is the second extension (\texttt{lid-n2v-rwpq}). 
The column ``I [\%]'' is the percentage improvement in $F_{1}$ of a better 
LID-elastic node2vec extension over the original node2vec.

\begin{table}[htb!]
\caption{Comparison of Node2Vec and LID-elastic Node2Vec embeddings.}
\small
\begin{center}
\begin{tabular}{lllllllll}
\noalign{\smallskip}\hline \noalign{\smallskip}
& \multicolumn{2}{l}{\texttt{n2v}} & \multicolumn{2}{l}{\texttt{lid-n2v-rw}} & \multicolumn{2}{l}{\texttt{lid-n2v-rwpq}} & \multicolumn{2}{l}{} \\
\noalign{\smallskip}\hline \noalign{\smallskip}
Graph       & $F_{1}$ & Dim. & $F_{1}$ & Dim. & $F_{1}$ & Dim. & Best & I[\%]  \\
\noalign{\smallskip}\hline \noalign{\smallskip}
BlogC & 0.247 & 50 & 0.363 & 100 & 0.2 & 100 & \texttt{rw} & 47.16 \\
DBLP & 0.403 & 25 & 0.441 & 25 & 0.531 & 50 & \texttt{rwpq} & 31.7 \\
GrQc & 0.563 & 10 & 0.707 & 25 & 0.657 & 50 & \texttt{rw} & 25.63 \\
CITESEER & 0.236 & 10 & 0.248 & 10 & 0.28 & 10 & \texttt{rwpq} & 18.69 \\
Zachary & 0.779 & 100 & 0.829 & 50 & 0.852 & 100 & \texttt{rwpq} & 9.41 \\
PUBMED & 0.394 & 50 & 0.431 & 50 & 0.42 & 50 & \texttt{rw} & 9.37 \\
CondMat & 0.653 & 25 & 0.696 & 50 & 0.702 & 50 & \texttt{rwpq} & 7.55 \\
AE photo & 0.495 & 50 & 0.52 & 50 & 0.487 & 50 & \texttt{rw} & 4.91 \\
cit-HepPh & 0.528 & 100 & 0.553 & 100 & 0.529 & 50 & \texttt{rw} & 4.8 \\
AE computers & 0.452 & 50 & 0.473 & 100 & 0.421 & 50 & \texttt{rw} & 4.67 \\
CORA & 0.572 & 25 & 0.595 & 50 & 0.59 & 50 & \texttt{rw} & 3.95 \\
cit-HepTh & 0.572 & 100 & 0.594 & 100 & 0.565 & 50 & \texttt{rw} & 3.9 \\
FB & 0.638 & 25 & 0.659 & 25 & 0.655 & 25 & \texttt{rw} & 3.22 \\
Les miserables & 0.81 & 100 & 0.799 & 100 & 0.832 & 200 & \texttt{rwpq} & 2.65 \\
HepPh & 0.708 & 25 & 0.724 & 50 & 0.7 & 25 & \texttt{rw} & 2.34 \\
AstroPh & 0.711 & 50 & 0.724 & 100 & 0.69 & 50 & \texttt{rw} & 1.78 \\
CORAML & 0.649 & 25 & 0.657 & 50 & 0.631 & 25 & \texttt{rw} & 1.28 \\
Florentine & 0.964 & 100 & 0.964 & 100 & 0.964 & 100 & all & 0 \\
\noalign{\smallskip}\hline \noalign{\smallskip}
\end{tabular}
\label{table_lid_n2v}
\end{center}
\end{table}

For Florentine families (the smallest graph in our experimental corpus) both LID-elastic node2vec extensions achieve 
the same $F_{1}$ score as node2vec. In all other cases at least one LID-elastic variant is better than node2vec. 
Both LID-elastic variants better preserve the graph structure than node2vec for 9 graphs 
(out of 18 in total). The \texttt{lid-n2v-rw} variant achieves the highest score $F_{1}$ for 12 graphs, 
while \texttt{lid-n2v-rwpq} wins on 5 graphs. Large improvements in $F_{1}$ scores are present for 4 graphs:  
\begin{itemize}
\item \texttt{lid-n2v-rw} significantly outperforms node2vec on BlogC and GrQc by increasing $F_{1}$ by 47.16\% and 25.63\%, respectively, and
\item \texttt{lid-n2v-rwpq} significantly outperforms node2vec on DBLP and CITESEER where $F_{1}$ is improved by 31.7\% and 18.69\%, respectively.
\end{itemize}
Considerable improvements in $F_{1}$ scores (approximately 5\% or higher) can be also observed for 5 graphs 
(Zachary, PUBMED, CondMat, AE photo, cit-HepPh and AE computers). Therefore, it can be concluded that our 
LID-elastic node2vec extensions are able to improve node2vec embeddings with respect to 
graph reconstruction errors.

\section{Conclusions and Future Work}

In this paper we have discussed the notion of local intrinsic dimensionality
in the context of graphs. Since graphs are dimensionless objects, 
existing LID models could be applied to graphs by computing LID estimators 
either on graph embeddings or on graph-based distances. 

Inspired by the fundamental connection between
the local intrinsic dimensionality and the discriminability of distance functions in
Euclidean spaces, we have proposed the NC-LID metric quantifying the
discriminability of the shortest path distance considering natural
communities of nodes in graphs. 
It has been shown that NC-LID exhibits consistent and stronger correlations to 
link reconstruction errors in node2vec embeddings than five centrality measures commonly used to quantify structural importance of nodes in complex networks. This result implies that NC-LID is a better choice for designing elastic graph embedding algorithms 
than standard node centrality measures.
Furthermore, we have evaluated two LID-elastic extensions
of the node2vec graph embedding algorithm in which hyperparameters are
personalized per node and adjusted according to their NC-LID values. Our experimental
evaluation of the proposed LID-elastic extensions on 18 real-world graphs
revealed that they are able to improve node2vec embeddings with respect to 
graph reconstruction errors.

The present research could be continued in several directions, some of them being theoretical and 
some having a more practical flavor. 
The first theoretical direction is to investigate
possibilities for designing LID-related metrics reflecting the discriminability of
graph-based distance functions considering expanding subgraph localities.
In the same way as NC-LID, such metrics could be exploited to personalize and adjust 
hyperparameters of graph embedding algorithms.
The second direction having a more applicative nature is to examine LID-elastic 
node2vec embeddings in the context of machine learning tasks on graph-structured data 
(e.g., node clustering, node classification and link prediction). Finally, 
the NC-LID measure (or its derivatives based on expanding subgraph localities) could be incorporated into 
the loss function of graph autoencoders and graph neural networks in order to obtain 
graph embeddings by LID-aware deep learning techniques.

\section*{Acknowledgements}

\noindent
This research is supported by the Science Fund of the Republic of Serbia, \#6518241, AI -- GRASP.
The authors are grateful to anonymous reviewers for their constructive comments and suggestions that
helped improve the quality of the paper.

\bibliography{refs}

\end{document}